\documentclass{article}
\usepackage[utf8]{inputenc}

\usepackage{enumitem}
\usepackage{natbib}
\usepackage{graphicx}
\usepackage{subfigure}

\DeclareFixedFont{\ttb}{T1}{txtt}{bx}{n}{8.8} 
\DeclareFixedFont{\ttm}{T1}{txtt}{m}{n}{8.8}  

\usepackage{color}
\definecolor{deepblue}{rgb}{0,0,0.5}
\definecolor{deepred}{rgb}{0.6,0,0}
\definecolor{deepgreen}{rgb}{0,0.5,0}
\definecolor{gray}{rgb}{.5, .5, .5}

\usepackage{listings}

\newcommand\pythonstyle{\lstset{
language=Python,
basicstyle=\ttm,
commentstyle=\ttm\color{gray},
otherkeywords={self},             
keywordstyle=\ttb\color{blue},
emph={approximator, policy,
      },          
emphstyle=\ttb\color{deepred},    
morekeywords={TorchApproximator, EpsGreedy, Atari, Core, DQN}, 
stringstyle=\color{deepgreen},
frame=tb,                         
showstringspaces=false            %
}}

\lstnewenvironment{python}[1][]
{
\pythonstyle
\lstset{#1}
}
{}

\title{MushroomRL: Simplifying Reinforcement Learning Research}
\author{Carlo D'Eramo$^{1}$, Davide Tateo$^{1}$, Andrea Bonarini$^{2}$,\\Marcello Restelli$^{2}$, Jan Peters$^{1,3}$\\ \small $^{1}$IAS Laboratory, TU Darmstadt, Darmstadt, Germany \\ \small $^{2}$DEIB, Politecnico di Milano, Milano, Italy \\ \small $^{3}$Max Planck Institute for Intelligent Systems, T\"{u}bingen, Germany}
\date{January 2020}

\begin{document}

\maketitle

\begin{abstract}
    MushroomRL is an open-source Python library developed to simplify the process of implementing and running Reinforcement Learning (RL) experiments. Compared to other available libraries, MushroomRL has been created with the purpose of providing a comprehensive and flexible framework to minimize the effort in implementing and testing novel RL methodologies. Indeed, the architecture of MushroomRL is built in such a way that every component of an RL problem is already provided, and most of the time users can only focus on the implementation of their own algorithms and experiments. The result is a library from which RL researchers can significantly benefit in the critical phase of the empirical analysis of their works. MushroomRL stable code, tutorials and documentation can be found at \texttt{https://github.com/MushroomRL/mushroom-rl}.
\end{abstract}

\section{Introduction}
The advantages of Reinforcement Learning (RL)~\citep{sutton1998reinforcement} methodologies are mostly shown in terms of empirical performance, especially in the recent years after the emergence of Deep RL~\citep{mnih2015human}. Indeed, in the vast majority of research papers, experimental evaluation really makes a difference between a successful and well cited work, and a mostly unknown one.
The need of evaluating algorithms comes together with the necessity of implementing and testing them in a quick and reliable way; thus, to address these problems, several RL libraries have been developed and are currently used by researchers. However, these libraries have heterogeneous drawbacks, e.g. they only focus on benchmarking and are not easy to extend; they do not have a sufficiently large number of already implemented algorithms.

We introduce MushroomRL, a RL Python library that has been developed to create a flexible, easy to understand, and comprehensive RL framework. MushroomRL comes with a strongly modular architecture that makes it easy to understand how each component is structured and how it interacts with other ones; moreover it provides an exhaustive list of RL and Deep RL methodologies that are ready to be used as baselines. In this paper, we provide an overview of the most famous RL libraries and briefly compare them with MushroomRL; then, we present the main ideas of our library and how they make MushroomRL powerful and unique; finally, we describe a complex use case to provide an overview of the way MushroomRL can be helpful in implementing and testing a novel complex RL algorithm.

\section{Related works}
The number of open-source RL libraries has significantly increased consequently to the success of Deep RL. \textit{OpenAI Baselines}~\citep{baselines} is one of the most famous examples of Deep RL libraries. It implements the majority of the most recent techniques and allows to test them on well-known RL problems through an interface with the benchmarking framework OpenAI Gym~\citep{gym}. The main recognized drawbacks of this library are its complex architecture and the difficulty in understanding the code, thus researchers are discouraged to extend it with novel functionalities and just use it for running available baselines. The more recent \textit{Stable Baselines}~\citep{stable-baselines} is a fork of OpenAI Baselines specifically made with the purpose of simplifying its architecture. However, despite the simplified interface, extending it with novel methodologies is still not straightforward. Another library is \textit{KerasRL}~\citep{plappert2016kerasrl} which is built on the well-known Deep Learning library \textit{Keras}. Unfortunately, this library is not maintained anymore. \textit{ChainerRL} is a Deep RL library based on the Deep Learning library \textit{Chainer}. Considering its structure and ideas, ChainerRL can be compared to Keras RL, but is still well maintained and documented. An example of a flexible RL library is \textit{Tensorforce}~\citep{tensorforce}, which is strongly based on Tensorflow. This library implements several Deep RL algorithms and its structure allows to easily test them on custom problems, besides the already available ones. Moreover, its modular architecture facilitate the process of extending the library with novel methodologies. Unfortunately, this library lacks of a complete documentation. Eventually, older RL libraries and interfaces have been proposed in the past, such as: \textit{RL-Glue}~\citep{tanner2009rl}, \textit{RLPy}, \textit{RLLab}~\citep{duan2016benchmarking}. However, these are not supporting most recent Deep RL techniques and moreover most of them are abandoned projects.

\section{Ideas and content}\label{S:ideas}
MushroomRL is developed with the explicit goal of addressing all the issues of similar RL libraries. The following are the main qualities of MushroomRL that make it a unique, yet powerful way of performing RL empirical research.
\paragraph{General purpose} The research on RL is not only about Deep RL. Shallow RL techniques (e.g. Q-Learning) are still important algorithms to consider, but most of RL libraries ignore them. Since there are no fundamental differences between shallow RL and Deep RL, MushroomRL adapts to heterogeneous learning tasks just focusing in modeling the interaction of an \textit{agent} with an \textit{environment}. This is achieved by a common interface which unifies a broad variety of RL techniques, such as: batch and online algorithms, episodic and infinite horizon tasks, on-policy and off-policy learning, shallow and Deep RL~(see Table \ref{T:algs}).
\paragraph{Lightweight}
MushroomRL is both user-friendly and flexible: only a high-level interface is exposed to the user, hiding low-level aspects. For instance, the user should not care about the implementation details to use a function regressor for different tasks, since they are hidden by a simple common interface. However, we leave the check of consistency constraints to the user, e.g. avoiding the use of a tabular algorithm for an environment with continuous state space. Minimal interfaces simplify the implementation of new algorithms, as there are no hard constraints in the prototypes.
\begin{table}[]
\centering
\begin{tabular}{c|p{5cm}|p{5cm}}
 & \multicolumn{1}{c|}{\textbf{Value-based}} & \multicolumn{1}{c}{\textbf{Policy-search}} \\
 \hline
 \textbf{RL} & Q-Learning, Double Q-Learning, Weighted Q-Learning, Speedy Q-Learning, R-Learning, SARSA, Expected SARSA, True Online SARSA-$\lambda$, FQI, LSPI & REINFORCE, GPOMDP, eNAC, RWR, PGPE, REPS, COPDAC-Q, Stochastic Actor-Critic \\
 \hline
 \textbf{Deep RL} & DQN, Double-DQN, Averaged-DQN, Categorical DQN & A2C, DDPG, TD3, SAC, TRPO, PPO
\end{tabular}\caption{Some of the available algorithms.}\label{T:algs}
\end{table}
\paragraph{Compatible}
Standard Python libraries useful for RL tasks have been adopted:
\begin{itemize}[noitemsep]
 \item \textit{Scientific calculus}: Numpy, Scipy;
 \item \textit{Basic ML}: Scikit-Learn;
 \item \textit{RL benchmark}: OpenAI Gym, DeepMind Control Suite~\citep{tassa2018deepmind}, Pybullet, MuJoCo~\citep{todorov2012mujoco}, ROS;
 \item \textit{Neural networks and GPU computation}: PyTorch.
\end{itemize}
MushroomRL provides an interface to these libraries, in order to integrate their functionalities in the framework, e.g. an interface for gym environments, support for regression with scikit-learn models.
\paragraph{Easy to use}
MushroomRL enables to develop and run experiments writing a minimal amount of code. In most of the tasks an experiment can be written in a few Python lines without the need of complex configuration files. The majority of the RL problems can be solved with experiments written following the structure of the library examples.

\section{Advanced use case: solving complex Reinforcement Learning problems}
The \texttt{examples} folder in MushroomRL provides several already implemented scripts that the user can run, and use as a reference to write their own. To show how easily a user can write and run complex experiments in MushroomRL, we consider the \texttt{atari\_dqn.py} script, which is used to run Deep Q-Network (DQN) algorithm~\citep{mnih2015human} on the Atari benchmark~\citep{bellemare2013arcade}.
This script generally follows the structure we recommend
\begin{python}
# Instantiation of problem, policy, and function approximator
mdp = Atari()
pi = EpsGreedy(...)
nn = TorchApproximator
# Instantiation of the agent (i.e. algorithm)
agent = DQN(approximator=nn, policy=pi, ...)
# Instantiation of the core
core = Core(agent, mdp)
# Learning and evaluation
core.learn(...)
core.evaluate(...)
\end{python}
\begin{figure}
\centering
\subfigure[Rendering of the game\label{F:breakout}]{\includegraphics[scale=.5]{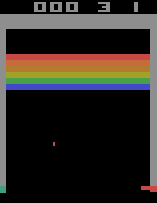}}
\hspace{.5cm}
\subfigure[Scores\label{F:scores}]
{\includegraphics[scale=.275]{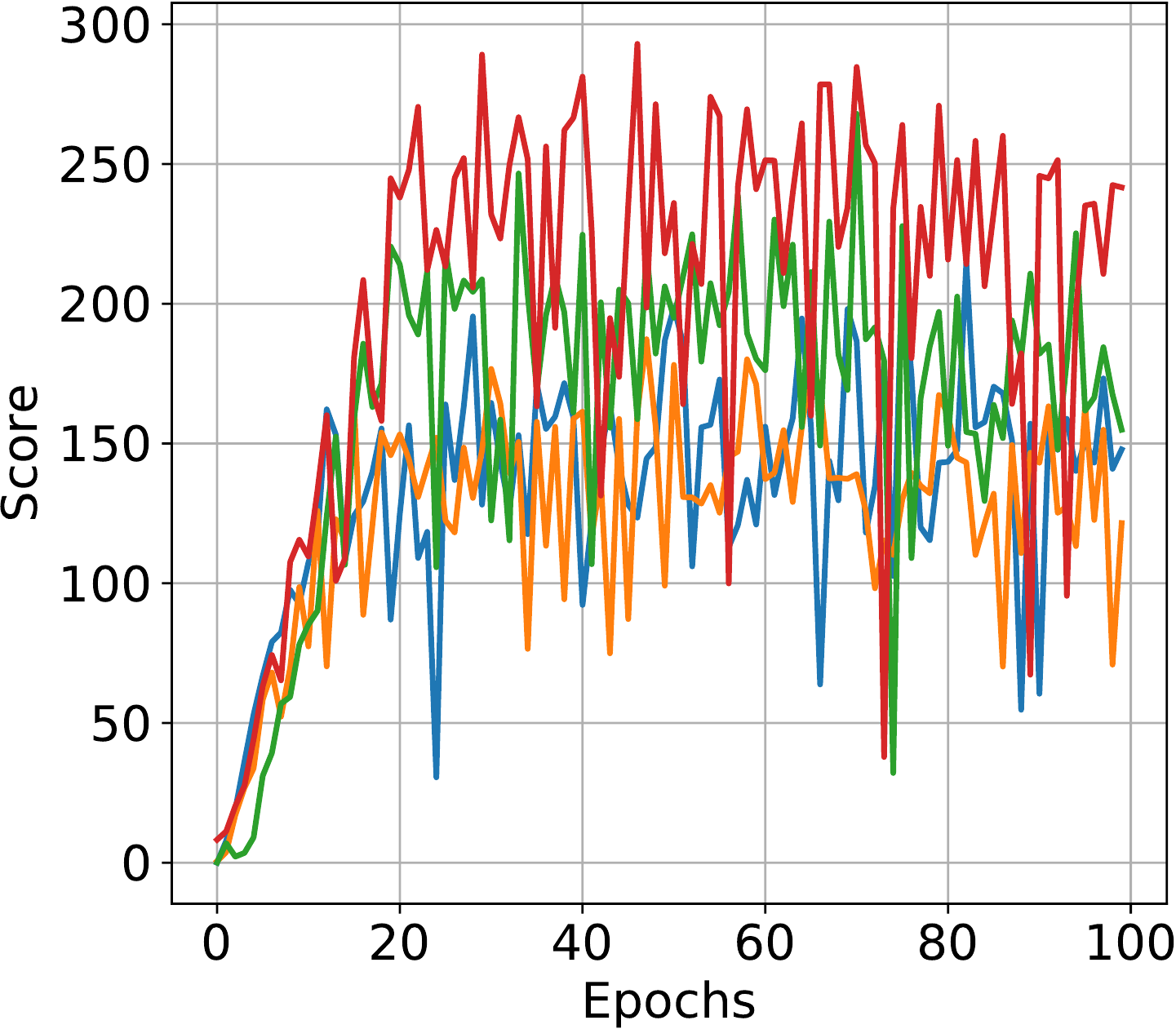}}
\hspace{.5cm}
\subfigure[Mean maximum $Q$-function\label{F:q}]
{\includegraphics[scale=.28]{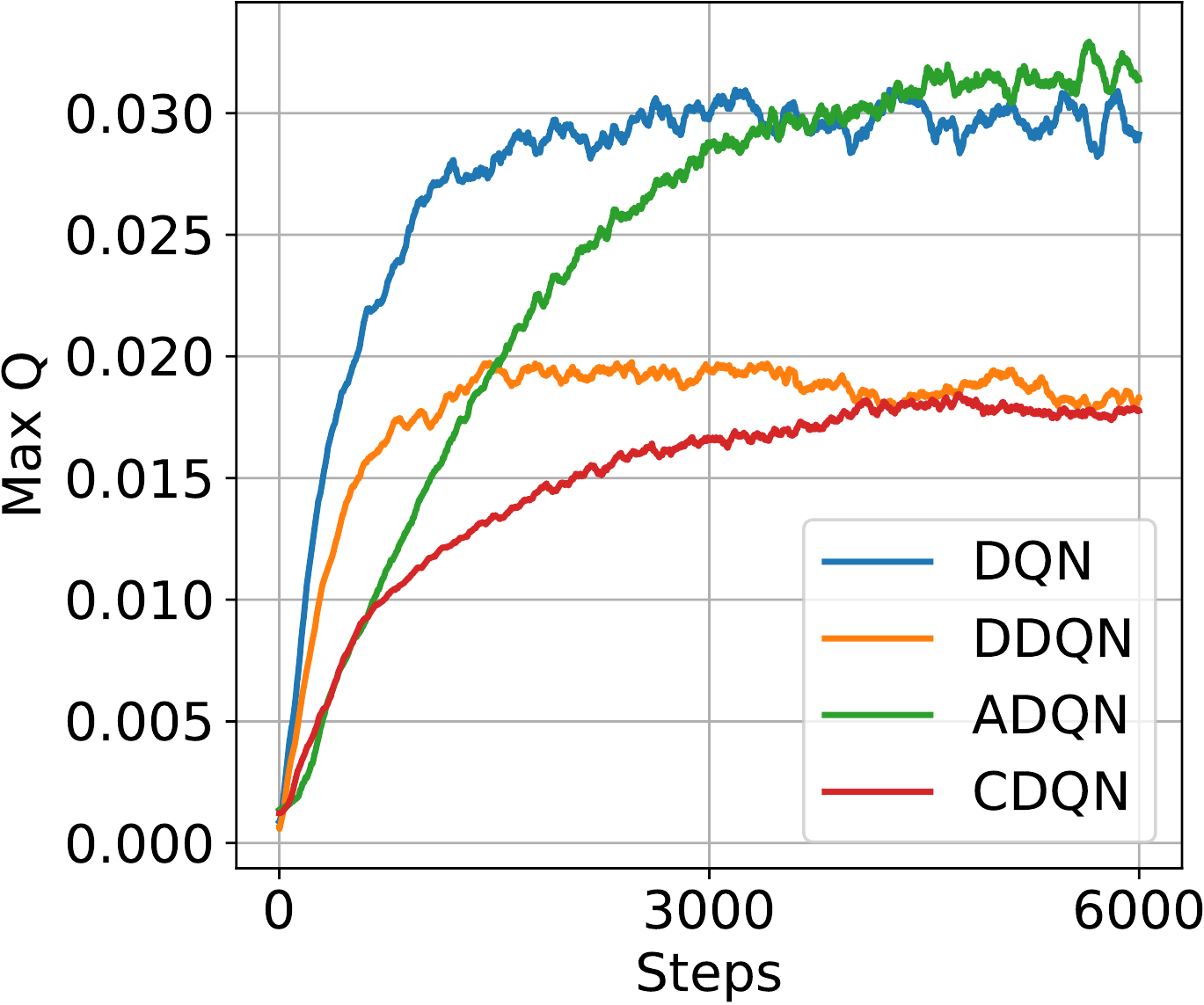}}
\caption{Rendering and some results on Breakout of DQN and some of its variants.}\label{F:exp}
\end{figure}
with some addition, e.g. the alternation between learn and evaluation phases as described in~\cite{mnih2015human}, and provide some useful functionalities that are easily integrated in the script, e.g. the use of the \texttt{argparse} library to parse input from the command line. Moreover, it saves the weights of the trained network, which can be loaded to initialize another network and evaluate it.
Furthermore, MushroomRL leaves the possibility to rely on other useful functionalities like plotting (e.g. using \texttt{matplotlib} library), or running multiple experiments in parallel (e.g. using \texttt{joblib} library). Figure~\ref{F:exp} shows some results obtained running \texttt{atari\_dqn.py} with DQN and four of its variants; note the high variance in the scores and the difference between the $Q$-values of DQN and DDQN, as described in literature. These results can be easily collected in MushroomRL through the use of the already provided callback functions, or new ones that the user can add. As stated in Section~\ref{S:ideas}, generally in MushroomRL only the low-level modules should not be modified by the user, while all the other ones can be customized and most of the functionalities provided by external libraries can be used, giving the user much freedom about the structure and analysis of his/her experiments.
In the case the user wants to add his/her own algorithm, he/she can simply add it in the \texttt{mushroom\_rl/algorithms/} and instantiate it in the script of the experiment.

\section{Conclusion}
We presented MushroomRL, an RL library to help researchers to easily develop their works and compare the results w.r.t. the majority of classical RL techniques and Deep RL approaches. Full documentation, installation instructions, and tutorials are available at \texttt{http://mushroomrl.readthedocs.io/en/latest}.

\newpage
\vskip 0.2in
\bibliographystyle{plain}
\bibliography{mushroom}

\end{document}